\documentclass{article}
\usepackage{spconf,amsmath,amssymb,graphicx,hyperref}
\usepackage{booktabs}
\usepackage{array}
\usepackage[table]{xcolor}
\usepackage{tipa}
\usepackage{tipx}

\title{Training-free L2-Accented Speech Generation}

\name{Thanathai Lertpetchpun$^{1}$, Yoonjeong Lee$^{1}$, Jihwan Lee$^{1}$, Tiantian Feng$^{1}$, Dani Byrd$^{2}$, Shrikanth Narayanan$^{1,2}$
\thanks{This work was supported by the Office of the Director of National Intelligence (ODNI), Intelligence Advanced Research Projects Activity (IARPA), via the ARTS Program under contract D2023-2308110001. The views and conclusions contained herein are those of the authors and should not be interpreted as necessarily representing the official policies, either expressed or implied, of ODNI, IARPA, or the U.S. Government. The U.S. Government is authorized to reproduce and distribute reprints for governmental purposes notwithstanding any copyright annotation therein.}}

\address{$^{1}$Signal Analysis and Interpretation Lab, University of Southern California, USA \\
  $^{2}$Department of Linguistics, University of Southern California, USA}

\begin{document}
\maketitle

\begin{abstract}
Accent plays a crucial role in speaker identity and broad utility of speech technologies. Existing accented Text-to-Speech (TTS) systems either require large-scale accented datasets or lack fine-grained phoneme-level controllability. The present work describes an accented TTS framework that combines coarse phonological rules with a pretrained multilingual TTS model, without any additional data of accented speech or target-accent speaker embeddings. Additionally, the trade-off between explicit phoneme-level duration alignment and natural language-specific timing is examined. Experimental results demonstrate effective accent shifts while maintaining speech quality and perceptual naturalness, showing that combining knowledge-based phonological transformations with native-language speaker embeddings can generate controllable L2-accented English.
\end{abstract}


\section{Introduction}

As the global \textit{lingua franca}, English has evolved into a decentralized linguistic landscape. Current estimates indicate that non-native (L2) speakers outnumber native (L1) speakers by a ratio of three to one \cite{crystal2003english}. This demographic reality is mirrored by a vast spectrum of phonetic variation in global English, ranging from regional L1 varieties—such as American, British, and Singaporean English \cite{wells1982accents}—to L2 varieties shaped by the cross-linguistic influence and distinct phonotactic patterns of a speaker's native language \cite{kachru1990world,schneider2007postcolonial}.


Despite this varied landscape, existing Text-to-Speech (TTS) systems have focused on a narrow subset of mainstream accents (e.g., North American or British English), often failing to represent the authentic speech patterns of the global majority~\cite{casanova2022yourtts, kim2021conditional, chen2025neural, chen2024vall}. Accented text-to-speech (TTS) models aim to control the accent of synthesized speech. A common approach is to fine-tune a pretrained TTS model on accented datasets~\cite{liu2024controllable, zhou2024accented}. However, collecting large-scale, high-quality L2-accented speech corpora remains costly and time-consuming. Although recent work \cite{inoue2025macst, lertpetchpun2026accent, onda2025prosodically} has begun to address this limitation, all of these lack fine-grained controllability over the phonetic mapping.


More recently, phonological rule-based transformations have been shown to enable controllable accent strength through phoneme modifications \cite{lertpetchpun2026quantifying}. This approach achieves fine-grained control at the phoneme level and effective accent manipulation between American and British English varieties. However, this framework assumes the availability of speaker embeddings that correspond to the target accent. Extending such an approach more generally to L2-accented English is not straightforward, as there is no idealized target speaker representation for accents arising from cross-linguistic influence. In this work, knowledge-based phonological transformations are combined with native-language (L1) speaker embeddings to generate controllable L2-accented English, bypassing the need for dedicated accented speech data or for target-accent speaker embeddings. 

We leverage a multistage pipeline grounded in multilingual TTS modeling. Many multilingual TTS systems are conditioned on both speaker embeddings and phoneme sequences~\cite{zhang2023speak, tran23d_interspeech, lu2025robust, fujita2024speech}. While speaker embeddings capture broad language and speaker identity characteristics~\cite{liu2023dse}, phoneme sequences provide a largely speaker-agnostic pronunciation representation. A phonological rule system is first applied to convert US phoneme sequences into target variants, such as Spanish-accented or Indian English. The modified phoneme sequences are then synthesized using a pretrained multilingual TTS model and a speaker embedding. Leveraging a multilingual backbone allows the system to draw on its cross-lingual prior knowledge, producing speech that reflects not only explicit segmental pronunciation alternations but also implicit suprasegmental characteristics such as rhythm and intonation.

Importantly, the proposed phonological transformations are \textit{not} intended to constitute exhaustive linguistic models of any particular L1-influenced English variety. \textbf{As a canonical `accent' is an idealized construct rather than a homogeneous entity for any given individual, attempting to map every fine-grained dialectal feature perfectly is not feasible, nor likely desirable, for TTS systems.} Instead, our rules serve as a controlled framework for investigating the interaction between explicit, coarse-grained phoneme-level modifications and the implicit accent information embedded in multilingual speaker embeddings. By capturing only the most salient phonological shifts, these rule sets enable systematic manipulation of accent strength, leaving the realization of finer-grained pronunciation to the pretrained TTS model. Our objective is to enable the controllability and generalization capabilities of a phonology-driven accent generation framework in a zero-shot setting.

In summary, we present a framework in which explicit phonological rules and multilingual speaker embeddings \textit{jointly} contribute to the resulting accent characteristics. Core contributions are: 1) a knowledge-driven framework for generating accented speech using a pretrained multilingual TTS model without requiring any accented training data; 2) fine-grained control over accent strength with no additional model training; 3) experimental validation of rhythmic variations influenced by speakers’ native languages. Speech samples are available\footnote{\url{https://sav-eng.github.io/jasa_el_supplement}}. Source code will be released upon acceptance.

\section{Methods}

\subsection{Phonological Rules}

We evaluate the proposed approach by focusing on two target varieties: Spanish-accented English (SP) and Indian English (IN). The phonological transformation rules are knowledge-based and systematically selected according to two criteria: (1) linguistically documented phonotactic and phonemic properties of the respective L1 systems, and (2) established systematic phonological disparities between US English and these L1 structures. Each implemented rule models a highly salient, well-documented cross-lingual interaction arising from L1 segmental transfer~\cite{Flege2006SecondLS, Haugen_1954, Flege_Bohn_2021, baese2022nature}.

\begin{table}[t]
\caption{Phonological transformation rules for shifts from US English to Spanish-accented and Indian English.}
\centering
\label{tab:rules}
\scriptsize
\setlength{\tabcolsep}{3pt}
\begin{tabular}{lll}
\toprule

Spanish-Accented Rules& US IPA & Spanish IPA \\
\midrule

1. Initial Consonant Substitution 
& \textipa{/v \texttheta} \textipa{\dh} \textipa{z \textdyoghlig/} 
& \textipa{/b s d s j/} \\

2. Spanish Rhoticity 
& \textipa{/\*r/} 
& \textipa{/r R/} \\

3. Epenthesis (s-clusters)
& \textipa{/sp st sk/} 
& \textipa{/esp est esk/} \\

4. Final Consonant Devoicing
& \textipa{/b d g z \textdyoghlig/} 
& \textipa{/p t k s \textteshlig/} \\

5. Vowel Simplification
& \textipa{/I U @ A 2 E 3 O/}
& \textipa{/i u a a a e e o/} \\

6. Monophthongization and Schwa
& \textipa{/eI oU @/} 
& \textipa{/e o a/} \\

\midrule

Indian English Rules& US IPA & Indian IPA \\
\midrule

1. Retroflexion of Stops and R
& \textipa{/t d \*r/} 
& \textipa{/\:t \:d \textlonglegr/} \\

2. Dentalization of Fricatives
& \textipa{/\texttheta} \textipa{D \textbeltl/} 
& \textipa{/\|[t \|[d l/} \\

3. Consonant Substitutions
& \textipa{/v Z/} 
& \textipa{/w z/} \\

4. Vowel Simplification
& \textipa{/I U \ae} \textipa{2} \textipa{\textinvscripta} \textipa{E 3 O/} 
& \textipa{/i u a @ a e e o/} \\

5. Monophthongization and Schwa
& \textipa{/eI oU @/} 
& \textipa{/e o a/} \\

\bottomrule
\vspace{-10mm}
\end{tabular}
\end{table}

\begin{table}[t]
\caption{Examples of applying phonological rules. The first column indicates the accented IPA transcription and the third column lists the phonological rules applied in the transformation.}
\scriptsize
\setlength{\tabcolsep}{3pt}
\centering
\label{tab:examples_rule}
\begin{tabular}{lll}
\toprule
IPA & Example & Rules \\
\midrule
\multicolumn{3}{l}{\textbf{This very tall teacher closed the big park.}} \\
\midrule
US & \textipa{\textbf{/D}Is \textbf{v}E\textbf{\*r}i \textbf{tO}l \textbf{t}i\textteshlig @\textbf{\*r} kl\textbf{O}z\textbf{d} \textbf{D}@ b\textbf{Ig} p\textbf{A}\*rk./} & \\
SP & \textipa{\textbf{/d}Is \textbf{b}E\textbf{R}i t\textbf{o}l ti\textteshlig @\textbf{R} kl\textbf{o}zd \textbf{d}@ bi\textbf{k} p\textbf{a}\*rk./} & 1,2,4,5,6 \\
IN & \textipa{/\textbf{\|[d}Is \textbf{w}E\textbf{R}i \textbf{\:to}l \textbf{\:t}i\textteshlig @\textbf{R} kl\textbf{o}z\textbf{\:d} \textbf{\|[d}@ b\textbf{i}g p\textbf{a}\*rk./} & 1,2,3,4,5 \\
\midrule

\multicolumn{3}{l}{\textbf{The little button was very broken.}} \\
\midrule
US & \textipa{/\textbf{D}@ l\textbf{I}R\textbf{@}l b\textbf{2}tn w\textbf{2z} \textbf{v}E\textbf{\*r}i b\*r\textbf{O}k@n./} \\
SP & \textipa{/\textbf{d}@ l\textbf{i}R\textbf{a}l b\textbf{a}tn w\textbf{as} \textbf{b}E\textbf{R}i b\*r\textbf{o}k@n./} & 2,5,6 \\
IN & \textipa{/\textbf{\|[d}@ lIR@l b\textbf{@}tn w\textbf{@}z \textbf{w}E\textbf{R}i b\*r\textbf{o}k@n./} & 1,2,3,4,5 \\
\midrule

\multicolumn{3}{l}{\textbf{Kate found three big red stones.}}\\
\midrule
US & \textipa{/k\textbf{eI}t faU\textbf{d} \textbf{\texttheta}\*ri b\textbf{Ig} \textbf{\*rE}d \textbf{st}On\textbf{z}./} \\
SP & \textipa{/k\textbf{e}t faU\textbf{t}  \textbf{s}\*ri b\textbf{ik} \*r\textbf{e}d \textbf{est}On\textbf{s}./} & 1,3,4,5,6 \\
IN &\textipa{/keIt faUd \textbf{\|[t}\textbf{}\*ri b\textbf{i}g \textbf{RE}d \textbf{s\:t}Onz./} & 1,3,4,5 \\
\bottomrule
\vspace{-7mm}
\end{tabular}
\end{table}

For Spanish-accented English, these rules capture structural substitutions driven by the phonotactic constraints and five-vowel architecture of the Spanish sound system~\cite{bradlow1995comparative}, including final-consonant devoicing and initial /s/-cluster epenthesis~\cite{flege1992production, carlisle1997modification}. For Indian English, the rules model the nativized, institutionalized features arising from ongoing contact with local languages (such as Hindi or Tamil), characterized for example by the retroflexion of alveolar stops and the systematic dentalization of English fricatives~\cite{pingali2009indian, Kachru, sirsa2013effects}.

Table~\ref{tab:rules} outlines the phonological transformation rules deployed. For each accented target variety, phoneme realizations occurring in the baseline General American English system (middle column) are deterministically mapped to their typical realizations in a target L1-influenced accent variety (right column). For example, the word \textit{three}, \textipa{/\texttheta \*ri/} is realized as \textipa{[s\*ri]} under the Spanish-accented ruleset by applying Rule~1 (Initial Consonant Substitution). Context-dependent allophonic variations and finer-grained phonetic nuances are intentionally excluded from the explicit rule mapping, as these subphonemic characteristics are left to be implicitly handled by the language-specific priors encoded within the target speaker embedding. 

Table~\ref{tab:examples_rule} provides sentence-level examples illustrating the cumulative application of these transformations. The first row displays the baseline US IPA transcription, while the subsequent rows present the transformed Spanish-accented and Indian English phoneme sequence outputs, with the final column tracking the specific rules applied. The transformations are executed across the entire input string according to a rule hierarchy established so as to avoid undesirable feeding and bleeding interactions among the rules. 


\vspace{-2mm}
\subsection{Accented Speech Generation}

\begin{figure}[t]
    \centering
    \vspace{-5mm}
    \includegraphics[width=0.8\linewidth]{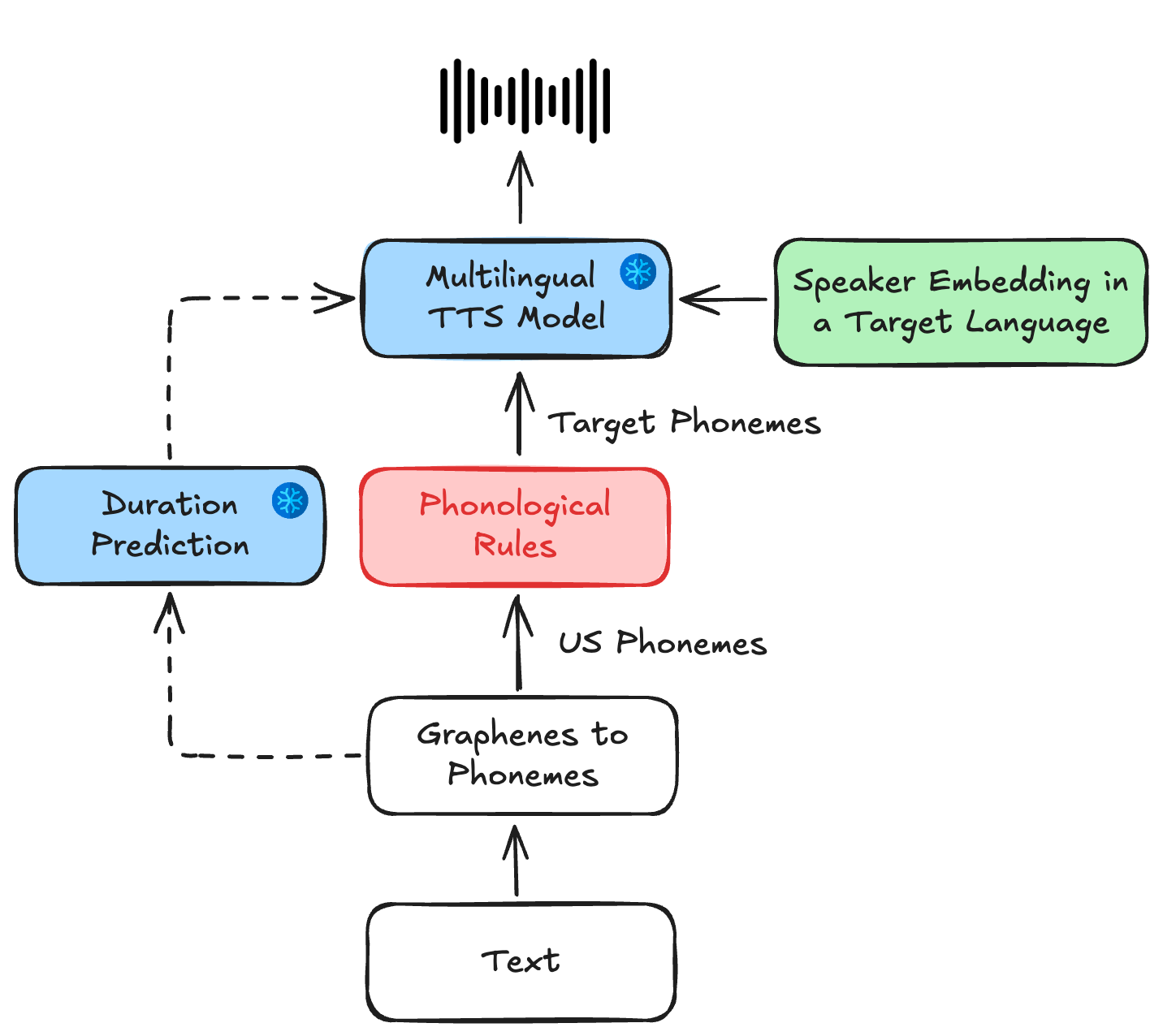}
    \caption{Synthesis Pipeline. The basic synthesis and evaluation pipeline is described in \cite{lertpetchpun2026quantifying}; the key difference is that the present work conditions the TTS model on a speaker embedding from the \textbf{target} language. In addition, the application of duration alignment is explicitly controlled between the American (US) phoneme sequence and the transformed target-accent phoneme sequence.}
    \vspace{-5mm}
    \label{fig:diagram}
\end{figure}

A pretrained multilingual TTS model is leveraged as the backbone of the synthesis pipeline. As illustrated in Fig.~\ref{fig:diagram}, the model accepts two decoupled inputs: a speaker embedding and a phoneme sequence. 

This approach exploits this implicit language encoding to synthesize L2-accented English without requiring explicit L2 training data. Specifically, the TTS model is conditioned on \textit{an L1 speaker embedding from the target source language} (e.g., native Spanish or Indian), while feeding it an English phoneme sequence that has been modified by the accent-specific phonological rules. In this way, the system forces the L1 speaker embedding to articulate English phonetic content. This cross-lingual interaction yields natural, intelligible, and controllable L2-accented English speech.

\vspace{-2mm}
\subsection{Rhythmic Differences}
We further investigate cross-linguistic rhythmic variations and their transfer effects in L2-accented speech. As pronunciation is strongly influenced by the prosodic system of a speaker's native language, timing and rhythmic transfer effects are expected to influence the resulting accented English realization. For instance, many major Indian languages, such as Hindi, are traditionally characterized as syllable-timed languages; in these varieties, sequential syllables tend to exhibit relatively regular duration and prominence, lacking severely reduced vowels. This contrasts with stress-timed languages such as English, which present significant vowel reduction in unstressed syllables~\cite{sirsa2013effects}. This means that varieties of Indian English reflect the regular temporal distribution of syllable cadence of local languages and therefore exhibit lesser durational contrast between stressed and unstressed syllables.

To isolate the impact of rhythmic transfer, duration modeling is manipulated within the synthesis pipeline. Specifically, two distinct conditions are evaluated: one in which accent-specific, native-like duration patterns are preserved by the model, and another in which phone durations are force-aligned to an American English timing baseline. 

\vspace{-2mm}
\section{Experiments}
\vspace{-2mm}
\subsection{Datasets and Experimental Setup}
We employ Kokoro-82M v0.19\footnote{\url{https://huggingface.co/hexgrad/Kokoro-82M}}, a pretrained multilingual TTS model, to generate all speech samples. The model space includes 20 American (US), 8 British (UK), 3 Spanish, and 4 Hindi speaker embeddings. 
For the evaluation of accent shifts and rhythmic variation (Tables~\ref{tab:main_results} and \ref{tab:rhythmic_variation}), we extract transcripts from the LibriTTS-R~\cite{libritts_r} \textit{train-clean-100} split, comprising 33k utterances. For the isolated rule ablation study (Table~\ref{tab:effect_rule}), we use transcripts from the LibriTTS-R \textit{test-clean} split, containing approximately 4.8k utterances.

\vspace{-2mm}
\subsection{Evaluation Metrics}
We evaluate the performance of the framework across two core dimensions: \textit{accent strength} and \textit{synthesis quality}, using a combination of objective and subjective metrics following \cite{lertpetchpun2026quantifying}.  (1) \textit{Accent Strength}: A narrow configuration of Vox-Profile \cite{feng2025vox} is employed, which outputs probabilities of each accent category and accent embedding including Romance and South Asia (proxies for Spanish and Indian) to evaluate the intended synthesized accented speech. (2) \textit{Synthesis Quality}: We measure the utility of the generated speech in terms of naturalness and intelligibility. Naturalness is evaluated using UTMOS \cite{saeki2022utmos}, an automatic predictor of Mean Opinion Score (MOS). The predicted score ranges from 1 (unnatural) to 5 (human-level naturalness). Intelligibility is evaluated using a pretrained Whisper-medium \cite{radford2023robust}. Word Error Rate (WER) and Character Error Rate (CER) are reported between the ASR-generated transcription and the ground-truth references.  

\textit{Subjective Evaluation}: To complement the objective assessments, a perceptual listening test was conducted evaluating: accent identification, perceived accent strength, and naturalness. Fourteen human evaluators, who are either native or fluent speakers of English, participated. The non-native fluent evaluators reside in the United States and represent varied L1 backgrounds (including Korean, Chinese, Thai, Japanese, Indian English, Norwegian, \& Danish). Most evaluators were based in the Los Angeles area, ensuring regular exposure to a wide range of English varieties, particularly Spanish-accented and Indian English. 70 samples were evaluated. For each trial, listeners performed a forced-choice accent identification task, followed by a prominence rating of the perceived accent on a 5-point Likert scale. Lastly, listeners rated the perceptual naturalness and resemblance to human speech using an identical 1-to-5 scale. 

\vspace{-2mm}
\section{Results and Discussion}
The framework is evaluated using three experimental setups: 1) \textit{Overall effectiveness of phonological rules}: Accentedness is evaluated as a function of classification probabilities and embedding similarities for American (US), Spanish (SP), and Indian (IN) accents. Additionally, UTMOS, WER, and CER are complementary and reported to evaluate naturalness and intelligibility. 2) \textit{Effect of duration alignment}: Results with and without duration alignment for each target accent are compared in order to analyze the impact of rhythmic control. 3) \textit{Rule ablation analysis}: We incrementally add individual phonological rules for each accent and measure their separate contribution to accentedness and overall performance.

\subsection{Phonological Rules and Accent Shifts}
Table~\ref{tab:main_results} demonstrates the overall effectiveness of the phonological rule application. When the accent rules are applied to a baseline (US) speaker embedding, the probability and embedding similarity drop substantially. This shift indicates a successful displacement away from an American English profile.

\begin{table}[t]
\caption{Overall effects of phonological rules on accentedness and speech quality. Accent probability and embedding similarity are reported for US and target accents, together with UTMOS, WER, and CER.}
\centering
\label{tab:main_results}
\scriptsize
\setlength{\tabcolsep}{2pt}
\begin{tabular}{lccccccc}
\toprule
& \multicolumn{2}{c}{Probability} 
& \multicolumn{2}{c}{Similarity} 
& \multicolumn{3}{c}{Speech Quality}
\\
\cmidrule(lr){2-3} \cmidrule(lr){4-5} \cmidrule(lr){6-8}
& \shortstack{US Prob.} & \shortstack{Tgt Prob.}
& \shortstack{US Sim.} & \shortstack{Tgt Sim.}
& UTMOS & WER & CER \\
\midrule
US Spk Emb. & \textbf{73.8} & -- & \textbf{0.71} & -- &
4.38 & 3.42 & 2.83 \\
+ SP Rules & \textbf{25.97} & -- & \textbf{0.50} & -- &
4.39 & 24.91 & 13.56 \\
+ IN Rules & \textbf{16.85} & -- & \textbf{0.42} & -- &
4.38 & 32.66 & 18.32 \\
\midrule
SP Spk Emb. & 26.6 & \textbf{23.7} &
0.73 & \textbf{0.47} &
3.73 & 7.99 & 5.64 \\
+ SP Rules & 2.97 & \textbf{51.59} &
0.38 & \textbf{0.60} &
3.87 & 13.13 & 7.79 \\
\midrule
IN Spk Emb. & 1.89 & \textbf{58.86} &
0.28 & \textbf{0.74} &
4.25 & 8.35 & 5.68 \\
+ IN Rules & 0.27 & \textbf{86.4} &
-0.008 & \textbf{0.75} &
4.16 & 24.15 & 14.78 \\
\bottomrule
\vspace{-8mm}
\end{tabular}
\end{table}

This shift is accompanied by an increase in downstream WER \& CER metrics.  Importantly, these elevated error rates reflect a known limitation in automatic speech assessment rather than an actual degradation of speech intelligibility as UTMOS scores remain highly stable across conditions. As standard ASR systems, including Whisper, are predominantly trained on Mainstream English variants, they inherently penalize systematic accent-induced phonetic variations \cite{peng2024evaluating}. For example, the Spanish-accented ruleset deterministically maps \textit{think} (\textipa{/TINk/}) to \textit{sink} (\textipa{/sINk/}). While an ASR system registers this substitution as a lexical error, it in fact represents the exact \textit{intended} behavior of our phonological transformation. Therefore, objective error rates (WER \& CER) should be interpreted with caution; they confirm that the phonological rules successfully alter the acoustic tokens away from the reference construct. 

When evaluating the primary pipeline configuration, i.e., applying the rulesets along with their corresponding target L1 speaker embeddings, a synergistic effect is observed\footnote{This behavior aligns closely with findings from previous frameworks \cite{lertpetchpun2026quantifying}. That work demonstrated that a target speaker embedding can occasionally attenuate or entirely overwrite explicit phoneme-level rule transformations, pulling the phonetic realization back toward its implicit language priors.  In our current cross-lingual setting, we observe a similar tension. For instance, when a target L1 embedding is forced to synthesize a heavily modified English sequence, certain fine-grained phone substitutions are smoothed out by the acoustic embedding's language-specific defaults. Rather than viewing this attenuation as a limitation, we exploit it as a stabilizing mechanism. The speaker embedding dynamically dampens potentially hyper-accented or unnatural rule combinations, ensuring the final output speech balances explicit accent salience with perceptual naturalness.}. For both Spanish and Indian configurations, introducing our explicit rules leads to a sharp increase in the respective target-accent probability and embedding similarity compared to using the L1 embeddings alone. This confirms the effectiveness of the proposed transformations in producing the desired accented English. Finally, across conditions, UTMOS scores remain highly stable, suggesting that perceptual naturalness is preserved without degrading quality in the modified English sequences.


\vspace{-2mm}
\subsection{Rhythmic Variation Effect}

Table~\ref{tab:rhythmic_variation} illustrates the perceptual and classifier effect of temporal control through phoneme-level duration alignment. By explicitly decoupling or imposing segment-level timing, we examine how timing influences accent classification. For the British (UK) and Indian (IN) conditions, removing explicit phoneme-level duration alignment results in a noticeable increase in target-accent probabilities. This confirms that preserving accent-specific phone duration patterns significantly strengthens accent perception. For the Spanish condition, although removing alignment does not increase the target-accent probability, it systematically decreases the posterior probabilities of non-target classes. This behavior indicates that language-specific timing variation contributes to accent differentiation, reducing the overlap between unrelated language variant groups. 

Importantly, our phonological rule modifications operate exclusively on phone identity while preserving the underlying segment duration from the US baseline. The variations observed in Table~\ref{tab:rhythmic_variation} serve to decouple segmental identity from temporal distribution, revealing  distinct accent-specific patterns. For Indian English, acoustic accent salience is driven by rhythmic cadence, allowing those phones to expand into their natural temporal pattering of being fully realized. Conversely, for Spanish, imposing the US timing baseline maintains or slightly enhances target probability. This suggests that the forced temporal compression of full-vowel substitutions into compressed, unstressed slots creates a highly salient acoustic pattern that the classifier associates with non-native accentedness, highlighting two distinct pathways via which explicit segmental rules interact with implicit suprasegmental modeling.

\begin{table}[t]
\vspace{2mm}
\caption{Effect of duration alignment on accentedness. Accent classification probabilities are reported for UK, Spanish (SP), and Indian (IN) conditions, with and without phoneme-level duration alignment.}
\label{tab:rhythmic_variation}
\centering
\scriptsize
\setlength{\tabcolsep}{3pt}
\begin{tabular}{lcccccc}
\toprule
 & \multicolumn{2}{c}{UK} & \multicolumn{2}{c}{SP} & \multicolumn{2}{c}{IN} \\
\cmidrule(lr){2-3} \cmidrule(lr){4-5} \cmidrule(lr){6-7}
& \shortstack{w/ Align.} & \shortstack{w/o Align.}
& \shortstack{w/ Align.} & \shortstack{w/o Align.}
& \shortstack{w/ Align.} & \shortstack{w/oAlign.} \\
\midrule
UK & 80.28 & 81.12 & 1.23  & 1.06  & 0.17  & 0.70 \\
SP & 6.50  & 3.90  & 51.59 & 50.86 & 3.28  & 2.73 \\
IN & 0.82  & 0.56  & 6.08  & 2.88  & 86.40 & 93.10 \\
\bottomrule
\vspace{-8mm}
\end{tabular}
\end{table}

\vspace{-3mm}
\subsection{Ablation Analysis of Phonological Rules}
Table~\ref{tab:effect_rule} demonstrates the effectiveness of each rule on transforming the accent. (The rules are ordered in the same order as shown in Table \ref{tab:rules}.) For the Spanish variety, Rule 5 (Vowel Simplification) is the single most influential driver of the accent shift. This aligns with empirical findings, as collapsing the complex English vowel space into a classic five-vowel system removes the vocalic properties utilized by native American-English classifiers. Conversely, for the Indian accent, Rule 1 (Retroflexion of Stops \& R) acts as the most prominent signature. These distinct results for rule hierarchy demonstrates that effective L2 synthesis depends on targeting the specific phonological pressures of the speaker's native (or matrix) language, rather than applying a generalized, uniform set of pronunciation shifts across all varieties of a language.


\begin{table}[t]
\caption{Effect of individual phonological rules on accentedness. Accent probability and embedding similarity are shown as each rule is added to the Spanish (SP) and Indian (IN) speaker embeddings.}
\scriptsize
\centering
\label{tab:effect_rule}
\begin{tabular}{lcccc}
\toprule
& US Prob. $\downarrow$ & SP Prob. $\uparrow$ & US Sim. $\downarrow$ & SP Sim. $\uparrow$ \\
\midrule
SP Spk Emb. & 32.46 & 22.86 & 0.70 & 0.32 \\
+ Rule 1 & 29.80 & 22.20 & 0.70 & 0.37 \\
+ Rule 2 & 21.13 & 23.00 & 0.66 & 0.32 \\
+ Rule 3 & 30.50 & 22.78 & 0.69 & 0.32 \\
+ Rule 4 & 28.47 & 25.22 & 0.69 & 0.35 \\
+ Best Rule (R5) & \textbf{9.82} & \textbf{25.72} &
\textbf{0.59} & \textbf{0.44} \\
+ Rule 6 & 24.25 & 20.12 & 0.65 & 0.33 \\
+ All Rules & \textbf{2.26} & \textbf{50.33} &
\textbf{0.43} & \textbf{0.58} \\
\midrule
& US Prob. $\downarrow$ & IN Prob. $\uparrow$ & US Sim. $\downarrow$ & IN Sim. $\uparrow$ \\
\midrule
IN Spk Emb. & 1.65 & 70.68 & 0.25 & 0.78 \\
+ Best Rule (R1) & \textbf{0.18} & \textbf{94.63} &
\textbf{0.06} & \textbf{0.83} \\
+ Rule 2 & 1.18 & 73.57 & 0.22 & 0.79 \\
+ Rule 3 & 1.30 & 72.93 & 0.22 & 0.79 \\
+ Rule 4 & 1.07 & 62.64 & 0.12 & 0.73 \\
+ Rule 5 & 1.04 & 79.18 & 0.20 & 0.79 \\
+ All Rules & \textbf{0.16} & \textbf{92.45} &
\textbf{-0.03} & \textbf{0.78} \\
\bottomrule
\vspace{-8mm}
\end{tabular}
\end{table}

\vspace{-2mm}
\subsection{Subjective Perceptual Evaluation}

\begin{table}[t]
\caption{Human evaluation: listener judgments of accent accuracy, perceived accent strength (for correctly identified samples), and naturalness.}
\scriptsize
\centering
\label{tab:results_human_eval}
\begin{tabular}{lccc}
\toprule
& Accuracy & Accent Strength & Naturalness \\
\midrule
US Spk Emb. & \textbf{1.000} & \textbf{4.04} & \textbf{4.09} \\
+ SP Rules & 0.386 & 2.59 & 2.81 \\
+ IN Rules & 0.229 & 2.88 & 2.73 \\
\midrule
SP Spk Emb. & 0.071 & 2.20 & \textbf{3.14} \\
+ SP Rules & \textbf{0.757} & \textbf{3.08} & 2.94 \\
\midrule
IN Spk Emb. & \textbf{0.786} & 3.31 & \textbf{3.47} \\
+ IN Rules & 0.757 & \textbf{3.94} & 3.14 \\
\bottomrule
\vspace{-8mm}
\end{tabular}
\end{table}

Human listening tests corroborate the objective findings, as shown in Table~\ref{tab:results_human_eval}. Applying the phonological rules to theUS speaker embedding noticeably shifts listener perception, successfully moving listeners away from identifying the speech as American. 

Crucially, when using the Spanish speaker embedding in isolation, most listeners overwhelmingly categorize the speech as American. This suggests that for certain non-native varieties within multilingual TTS, the speaker embedding alone is insufficient to signal the target accent. In contrast, incorporating the explicit Spanish phonological rules resolves this limitation. For the Indian accent condition, the baseline speaker embedding already encodes distinguishable accent features. Applying the explicit Indian English phonological rules maintains this high identification rate while intensifying the perceived accent strength. Notably, across all experimental conditions, human naturalness ratings remain highly stable, near or above a score of 3.0 (“Moderately Natural”). This demonstrates that manipulating salient structural accent characteristics does not degrade overall perceptual speech quality.


\vspace{-3mm}
\section{Conclusion}
In this work, we propose a training-free linguistically-driven framework for generating controllable L2-accented English speech. Rather than attempting to exhaustively replicate every hyper-local dialectal feature, which would treat accent as a monolithic, static entity, our approach targets a small set of salient, coarse-grained phonological shifts and unites this with native speaker embeddings in a multilingual TTS model. Objective metrics and subjective human evaluations validate the effectiveness of this framework, demonstrating reliable accent transformations for both Spanish-accented and Indian varieties of English while maintaining high speech quality and naturalness. Furthermore, our analysis of decoupled phone durations highlights the foundational role that suprasegmental rhythmic patterns play in accent salience and accent class differentiation. In sum, we demonstrate that combining knowledge-based phonological transformations with native-language (L1) speaker embeddings can generate controllable L2-accented English.


\bibliographystyle{IEEEbib}
\bibliography{refs}

\end{document}